\documentclass[preprint,12pt,authoryear]{elsarticle}
\usepackage{amsmath,amsfonts}
\usepackage{algorithmic}
\usepackage{algorithm}
\usepackage{array}
\usepackage[caption=false,font=normalsize,labelfont=sf,textfont=sf]{subfig}
\usepackage{textcomp}
\usepackage{stfloats}
\usepackage{url}
\usepackage{verbatim}
\usepackage{graphicx}
\usepackage{cite}
\usepackage{todonotes}
\usepackage[nocomma]{optidef}
\usepackage{amssymb}
\usepackage{textcomp}
\usepackage{natbib}
\usepackage[colorlinks=true]{hyperref}
\usepackage{multirow}

\begin{document}

\begin{frontmatter}
	
\title{Time-Optimal Path Tracking for Cooperative Manipulators: A Convex Optimization Approach\tnoteref{t1}}
\tnotetext[t1]{This work was supported in part by the Swedish Department of Education within the strategic research environment ELLIIT and in part by the Vinnova Competence Center LINK-SIC at Linköping University.}

\author[fn1]{Hamed Haghshenas\corref{cor1}}
\ead{hamed.haghshenas@liu.se}

\author[fn1]{Anders Hansson}
\ead{anders.g.hansson@liu.se}

\author[fn2]{Mikael Norrl{\"o}f}
\ead{mikael.norrlof@se.abb.com}

\cortext[cor1]{Corresponding author}

\fntext[fn1]{Hamed Haghshenas and Anders Hansson are with the Division of Automatic Control, Department of Electrical Engineering, Link{\"o}ping University, 581-83 Link{\"o}ping, Sweden.}
\fntext[fn2]{Mikael Norrl{\"o}f is with ABB Robotics Division, 721-36 V{\"a}ster{\aa}s, Sweden.}

\begin{abstract}
This paper studies the time-optimal path tracking problem for a team of cooperating robotic manipulators carrying an object. 
Considering the problem for rigidly grasped objects, we show that it can be cast as a convex optimization problem and solved efficiently with a guarantee of optimality.
When formulating the problem, we avoid using a particular wrench distribution and exploit the full actuation available to the system.
Then, we consider the problem for grasps using frictional forces and show that this problem also, under a force-closure grasp assumption, can be formulated as a convex optimization problem and solved efficiently and to optimality. To ensure a firm grasp, internal forces have been taken into account in this approach. 
\end{abstract}


\end{frontmatter}


\section{Introduction}
Robotic systems are nowadays the key technology in a wide range of application domains, from construction, manufacturing and agriculture to search and rescue, and service robotics. The increasing demand for performance of robotic systems is often met by using multiple robots for a specific task. A team of cooperating robots outperforms the functionality of a single robot; like a human using two arms has an advantage over one using only one arm. 
When a task conducted by multiple robots involves manipulation of an object, the multi-robot system is said to perform a \textit{cooperative manipulation} task.
Cooperative manipulation is an important capability for extending the domain of robotic applications. Typical examples include industrial manipulators manipulating large or heavy objects as well as transportation tasks conducted by multiple robots. 
One particular subject that most applications have in common is path tracking. Path tracking is the second stage of the so-called decoupled approach \citep{choset2005principles, lavalle2006planning}; an approach for solving motion planning problems. The first stage of the decoupled approach, known as path planning, determines a path while taking geometric aspects of the task and the environment into account, whereas the path tracking stage is concerned with the dynamic aspects of the robot and the task. In many robotic applications, the path is determined by the task and its specifications. Hence, assuming that a desired geometric path is given, the path tracking problem can be studied independently and is of value on its own.

For a wide range of robotic applications, it is desired to minimize the execution time of a task or some other criteria such as the energy consumed during the motion.
For example, robotic manipulators performing a variety of tasks are included in almost every production line, and there is an obvious relation between the execution time of the tasks and productivity.
Hence, time-optimal and in general optimal motion planning is of great significance for robotic systems. By optimizing the robots' motions while taking their dynamics into account, robots can fully exploit their capabilities.

Motivated by these observations, this paper studies the time-optimal path tracking problem for a team of cooperating robotic manipulators. 
We consider a scenario where an object is grasped by multiple manipulators.
Time-optimal cooperative path tracking addresses the problem of moving the object along a predefined geometric path in minimum time, which requires realizing a velocity as high as possible along the path while satisfying the imposed constraints on the motion.
Note that solving this problem gives the nominal trajectories. For online path tracking, the nominal, feedforward trajectories are combined with feedback in a control architecture such that robustness to uncertainties in model or environment is achieved (see e.g., \citet{olofsson2017path}). Combining the feedforward trajectories with a feedback is beyond the scope of this work and our focus is on solving the time-optimal path tracking problem. 
Most of the works on the area of time-optimal path tracking are focused on a single manipulator. Contrary to the case of a single manipulator, a given time parameterization of the path in the case of cooperative manipulators may correspond to infinitely many possible torque profiles. One possible way to deal with this redundancy is to consider a predefined load distribution. It is however clear that this would give rise to slower trajectories than what can be obtained using the full actuation available to the system. Our goal in this paper is to extend the time-optimal path tracking to the case of cooperative manipulators so that it can optimally handle the actuation redundancy. 

There are different methods for solving the time-optimal path tracking problem, most of which exploit the fact that the motion along a predetermined path can be described by a single parameter, called path coordinate and denoted by $s$, and its time derivatives \citep{shin1985minimum, bobrow1985time}. 
These approaches can basically be divided into three groups: numerical integration, dynamic programming, and convex optimization. 

The numerical integration approach \citep{shin1985minimum, bobrow1985time, pfeiffer1987concept, constantinescu2000smooth} is based on the Pontryagin maximum principle: the time-optimal velocity profile in the $(s,\dot{s})$ plane is known to be bang--bang and can be obtained by integrating successively the maximum and minimum accelerations $\ddot{s}$. 
This requires finding the switch points between accelerating and decelerating segments, which constitutes a major implementation difficulty as well as the main cause of failure \citep{slotine1988improving,shiller1992computation}. 
This is because of the robustness issues associated with the so-called dynamic singularities which occur when the maximum velocity curve is nondifferentiable (see \citet{pham2014general}).
Furthermore, finding the switch points and computing the maximum velocity curve become computationally expensive for complex constraints with many inequalities \citep{pham2019critically}.
This approach is theoretically faster than the other two approaches, since it exploits the bang-bang structure of the problem. 
A limitation of this approach is that it is restricted to the time-optimal case.
An algorithm for handling of dynamic singularities for robots subject to joint velocity and acceleration constraints is presented in \citet{kunz2012time}. In \citet{pham2014general}, a more general approach that can account for dynamic singularities resulting from second-order constraints together with an open-source implementation of the algorithm in C\texttt{++}/Python is presented. In \citet{shen2018complete}, a complete and time-optimal algorithm based on numerical integration is proposed.
In \citet{pham2017structure}, an algorithm that solely treats jerk limits is proposed. 
The numerical integration approach is extended to the case of redundantly actuated systems in \citet{pham2015time}. 
An object manipulated by several manipulators can be considered as a redundantly actuated system.
In their approach, they use the equations of motion of an open-chain system obtained by cutting the closed-chain system at some joints. This requires obtaining the dynamics of the open-chain system where the chain is cut. If the dynamic models of the manipulators are already available, obtaining the dynamics of another open-chain system can be an extra task. 
Also, to compute the optimal accelerations at the singular points, one cannot use the analytical value suggested in \citet{pham2014general} and has to search numerically for an approximate optimal acceleration. Furthermore, after obtaining the optimal parameterization, to obtain the torques at each actuated joint another optimization problem must be solved whose objective is specified by the user. 
Thus, the approach proposed in \citet{pham2015time} does not use the full actuation available to the system.
Compared to that work, our approach exploits the full actuation available to the manipulators. Also, it is straightforward to impose constraints on the wrenches acting on the object in our approach, whereas in \citet{pham2015time} it is not clear how to do this.
Time-optimal path tracking for cooperative multi-manipulator systems was independently studied in \citet{bobrow1990minimum} and \citet{moon1991time} based on numerical integration, where linear programming was used to find the maximum acceleration and deceleration at a given point in the $(s,\dot{s})$ plane. However, since the algebraic form of extreme accelerations is not available from linear programming, they are not able to develop a systematic search scheme to find the switching points, and their proposed procedures give an approximation to the time-optimal trajectory.

In the dynamic programming approach, the problem is solved by dividing the $(s,\dot{s})$ plane into a grid and subsequently using dynamic programming (see \citet{shin1986dynamic, oberherber2015successive}). One advantage of this method is the ability to account for a general form of the objective functions and constraints. 
The main disadvantage of this approach is the high computational cost due to the need for solving a problem with a large number of variables.
In \citet{kaserer2018nearly}, an algorithm which is capable of taking limits on joint jerks and torque rates into account is presented. Additionally, viscous friction is included in the dynamic model of the manipulator, which has been made possible by utilizing the idea of dynamic programming.


The convex optimization approach converts the original problem, which turns out to be an infinite-dimensional optimization problem, into a finite-dimensional one through discretization of the $s$-axis.
One of the early works using this approach is \citet{verscheure2009time}, where it is shown that by exploiting a nonlinear change of variables, the time-optimal path tracking problem can be cast as a convex optimization problem and subsequently as a second-order cone program (SOCP). This allows for efficiently solving the problem by utilizing the wide variety of algorithms and software developed for convex optimization \citep{boyd2004convex}. Furthermore, the convexity of the problem guarantees that any locally optimal solution is also a globally optimal one. 
This approach has also the advantage that other objective functions such as energy or torque rate, can be incorporated.
It must be noted that the change of variables used in \citet{verscheure2009time} has been known since at least 1985 (see e.g., \citet{bobrow1985time, dubowsky1986time, pfeiffer1987concept}), although the resulting convexity of the problem was not noted back then.
The main disadvantage of the convex optimization approach is computational cost, which can be an order of magnitude greater than numerical integration-based approaches.
To reduce computational cost, \citet{hauser2014fast} proposes an approach based on sequential linear programming. Further, \citet{nagy2018lp} shows that using a special discretization scheme, the time-optimal velocity profile can be obtained by linear programming with the benefit of lower computation cost with respect to convex solvers.

Because of the advantages that the convex formulation offers, much work has been conducted to extend the range of applicability of these ideas to other problems such as path tracking for different types of vehicles as in \citet{lipp2014minimum}, or to include new types of constraints in the robot path tracking problem \citep{ardeshiri2011convex, reynoso2013time, debrouwere2013time}. Similarly as in \citet{pfeiffer1987concept}, viscous friction which renders the problem nonconvex is not considered in the robot model in \citet{verscheure2009time}. In this context, some works such as \citet{ardeshiri2011convex} and \citet{debrouwere2013time} have addressed the incorporation of the constraints that destroy the convexity of the problem. The work \citet{ardeshiri2011convex} incorporates speed dependent constraints into the problem and replaces the nonconvex constraints with their convex approximations to preserve the convexity of the problem, whereas the work \citet{debrouwere2013time} considers several robotic applications that result in nonconvex problems and employs sequential convex programming to solve the corresponding problem by writing the nonconvex constraints as differences of convex functions.  
In \citet{zhang2016time}, change rate of torque and voltage of the DC motor are incorporated. By introducing some new variables which are tight approximation of change rate of torque and motor voltage, the original nonconvex problem is formulated as an approximate convex optimization problem.
The work by \citet{cao2016time} formulates the time-optimal path tracking problem for a dual-robot system as a nonconvex problem and solves an approximate problem using sequential convex programming.
The change of variables introduced in \citet{verscheure2009time} is also used in other studies. In \citet{steinhauser2018efficient}, a two-step iterative learning algorithm for path tracking problem of a robotic manipulator is proposed which compensates for a possible model-plant mismatch and improves the tracking performance. 

From the mentioned approaches for solving the time-optimal path tracking problem, only the numerical integration approach has been applied to cooperative manipulators. As mentioned earlier, the works that use this method for cooperative manipulators either do not use the full actuation available to the system \citep{pham2015time,moon1991time} or are not able to develop a systematic search for finding the switch points \citep{moon1991time}. Additionally, this approach suffers from robustness issues associated with the dynamic singularities and is limited to the time-optimal case. Because of the advantages that the convex optimization approach offers, in this paper we use this method to solve the time-optimal path tracking problem for a team of cooperating manipulators. With the availability and maturity of SOCP solvers, this approach reliably and robustly solve the problem of interest. With this approach, it is also possible to incorporate objective functions other than time into the problem.
First, we consider a scenario where an object is rigidly grasped by multiple manipulators. The rigid grasping assumption implies that the manipulators can apply forces and moments along all directions to the object. By exploiting the mentioned change of variables we show that this problem can be formulated as a convex optimization problem and subsequently as an SOCP. 
Due to multiple manipulators being involved in the manipulation task, there exists an infinite number of forces/moments exerted by the end-effectors that result in the same desired force/moment on the centre of mass of the object. 
This redundancy is fully exploited in our formulation. Others (e.g., \citet{moon1991time}) use specific force/moment distributions, also called wrench distribution, which result in suboptimal solutions.
We then relax the assumption on the rigid grasping and consider contacts with friction. 
These types of contacts complicate the problem as each contact may only apply forces/moments that respect friction cone constraints to prevent slip, instead of arbitrary forces/moments associated with rigid contacts. 
We show that this problem also, under a force-closure grasp assumption, can be formulated as a convex optimization problem.

In summary, the following are the contributions of this paper:
\begin{enumerate}
	\item The time-optimal path tracking problem for a team of cooperative manipulators rigidly grasping an object is formulated as a convex optimization problem.
	\item The results from the first contribution are extended to consider contacts with friction.
    \item Our approach optimally handles the actuation redundancy and uses the full actuation available to the system.
\end{enumerate}

In \citet{haghshenas2019convex}, we have presented a preliminary study of the time-optimal path tracking problem for a particular setup comprised of two two-link planar manipulators with non-actuated end-effectors rigidly grasping a bar. Compared to that work where only two manipulators are involved and the object is a bar grasped at the two end points, in this paper we consider a general scenario with $N$ generic manipulators and a generic object. Furthermore, a particular wrench distribution was assumed in \citet{haghshenas2019convex}, whereas here we avoid using any wrench distribution. 

The rest of this paper is organized as follows.
Section \ref{sec:preliminary} presents some known results on time-optimal path tracking problem for a single manipulator.
Section \ref{sec:probleStatement} states the problem of interest. Section \ref{sec:modeling} presents the modeling of the coupled kinematics and the dynamics of the manipulators and the object. Section \ref{sec:mainResult} studies the time-optimal cooperative path tracking problem with rigid contacts.
Section \ref{sec:friction} extends the results in Section \ref{sec:mainResult} to consider contacts with friction.
Section \ref{sec:simulation} contains some simulation results.
Section \ref{sec:conclusion} summarizes our findings and presents ideas for future work.


\section{Preliminaries}\label{sec:preliminary}

Here, we collect some known results about the time-optimal path tracking problem for a single manipulator that will be used in the rest of the paper. 


Manipulator motion along a prescribed trajectory can be written as a function of a single parameter, $s$, either in task space \citep{bobrow1985time} or in joint space \citep{shin1985minimum}.
Given a prescribed geometric path $q(s)$ in joint space, the joint velocities and accelerations can, using the chain rule, be written as
\begin{subequations}\label{eq_chain_single}
	\begin{align} 
		\dot{q}(s) &= {q}'(s) \dot{s}, \label{eq_qdot_single} \\
		\ddot{q}(s) &= {q}'(s) \ddot{s} + {q}''(s) \dot{s}^2, \label{eq_qdoubledot_single}
	\end{align}
\end{subequations}
where ${q}'(s) = \partial q(s)/\partial s$, ${q}''(s) = \partial^2q(s)/\partial s^2$, $\dot{s} = ds/dt$ and $\ddot{s} = d^2s/dt^2$. It is shown in \citet{verscheure2009time} that the time-optimal path tracking problem for a single robotic manipulator subject to lower and upper bounds on the torques, can be cast as the convex optimization problem
\begin{mini!}[2]
	{
	}{\int_{0}^{1}\frac{1}{\sqrt{b(s)}}ds}
	{\label{op:convex}}
	{\label{op:obj}}{}
	\addConstraint{\tau(s)}{=m(s)a(s)+c(s)b(s)+g(s), \label{op:convex:tau}}
	\addConstraint{b(0)}{=\dot{s}_0^2,}
	\addConstraint{b(1)}{=\dot{s}_T^2,}
	\addConstraint{b(s)}{\geq 0,}
	\addConstraint{{b}'(s)}{=2a(s),}
	\addConstraint{\underline{\tau}(s) \leq \tau(s) \leq \overline{\tau}(s),}{\label{op:convex:bounds}}
	\addConstraint{\forall s \in [0,1],}{}
\end{mini!}
where the optimization variables in addition to joint torques, $\tau(s)$, are the acceleration and square of the speed along the path coordinate, i.e.,
\begin{subequations}\label{eq:ab}
	\begin{align}
		a(s) &= \ddot{s}(t), \\
		b(s) &= \dot{s}(t)^2.
	\end{align} 
\end{subequations}
The key behind this result is to use the acceleration as a free variable, which can be accomplished by representing the joint velocities and accelerations as \eqref{eq_chain_single}, and substituting them in the equations of motion. 
The constraint in \eqref{op:convex:tau} shows the equations of motion of a single manipulator after this substitution and using \eqref{eq:ab} as optimization variables. The change of integration variable from time $t$ to $s$ allows to write the objective function, i.e. the duration of the motion, as \eqref{op:obj}.
In the above optimization problem, $\underline{\tau}$ and $\overline{\tau}$ are the lower and upper bounds on the joint torques, respectively, and can be functions of $s$, and $\dot{s}_0$ and $\dot{s}_T$ are, respectively, the initial and final velocities along the path, usually chosen to be $0$. In this problem, the inequalities are interpreted as component-wise inequalities.
The optimization problem in \eqref{op:convex} is convex since the objective function is convex in $b(s)$, which follows from the fact that integration preserves convexity and that $1/\sqrt{b(s)}$ is convex in $b(s)$, and since all the constraints are linear in the optimization variables. 
Extensions of the problem to incorporate other objective functions and constraints that preserve convexity can be found in \citet{verscheure2009time}. In particular, it is shown that symmetric lower and upper bounds on the joint velocities can be translated into upper bounds on $b(s)$. We denote this upper bound by $\overline{b}(s)$, and the corresponding constraint will be incorporated into the forthcoming optimization problems.

The convex optimization problem in \eqref{op:convex} is an infinite dimensional problem with infinitely many optimization variables and constraints. 
Therefore, the direct transcription method is employed and the problem in \eqref{op:convex} is formulated as a large sparse optimization problem in \citet{verscheure2009time}. The resulting problem is finally transformed into an SOCP, which can be solved efficiently using dedicated solvers, such as MOSEK \citep{mosek}. The reader is referred to \citet{verscheure2009time} for more details. 


\section{Problem statement}\label{sec:probleStatement}
\begin{figure}[!t]
\centering
\includegraphics[width=3.5in]{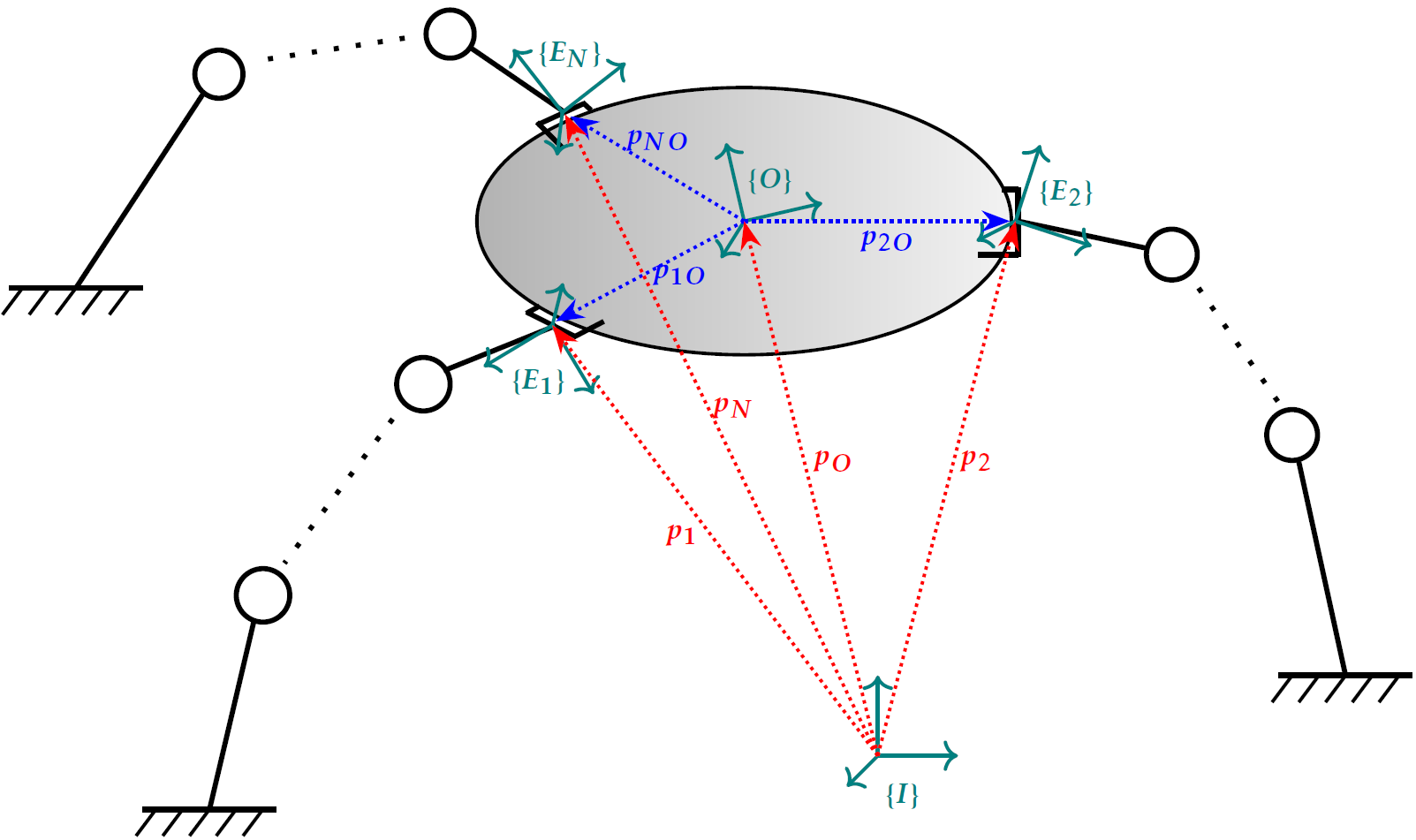}
\caption{Three robotic manipulators rigidly grasping an object.}
\label{fig:graspingN}
\end{figure}

This section addresses the problem of time-optimal path tracking for multiple cooperative manipulators rigidly grasping an object.
Consider $N$ fully actuated robotic manipulators, indexed by the set $\mathcal{N} = \{1, \ldots, N\}$, rigidly grasping an object as in \figurename~\ref{fig:graspingN}. The rigidity assumption implies that the manipulators can exert forces and moments along all directions of the object. The objective is to minimize the traversal time required to move the object with a desired orientation along a prescribed geometric path subject to constraints on the joint torques and velocities. We assume that the prescribed geometric path is given for the centre of mass of the object, and that both the path and the object's orientation are given as functions of a scalar path coordinate $s$. 
Without loss of generality, assume that the trajectory starts at $t=0$, ends at $t=T$, and that $s(0)=0 \leq s(t) \leq s(T)=1$. Furthermore, it is assumed that $\dot{s}(t) \geq 0, \forall t \in [0,T]$. In other words, we always move forward along the path. 

Throughout the paper, all quantities are expressed with respect to an inertial frame of reference $\{I\}$ whenever the reference frame is not explicitly indicated by a superscript. 
Let $\{O\}$ and $\{E_i\}$ denote the reference frames attached to the object at the centre of mass and the $i$th end-effector, respectively.
Let $\phi_{\scriptscriptstyle O}(s) = [\varphi_{\scriptscriptstyle O}(s), \vartheta_{\scriptscriptstyle O}(s), \psi_{\scriptscriptstyle O}(s)]^T$ be the object's orientation in terms of a triplet of Euler angles. 

In the following section, we present the modeling of the coupled kinematics and the dynamics of the manipulators and the object.

\section{Kinematic and Dynamic Models}\label{sec:modeling}
\subsection{Kinematics}
Let $p_{\scriptscriptstyle O} \in \mathbb{R}^{3}$ be the position of the object's centre of mass. Let $p_i \in \mathbb{R}^{3}$ and $\phi_i$ be the position and orientation of the $i$th end-effector, respectively. In view of \figurename~\ref{fig:graspingN}, one can see that the pose of the $i$th end-effector and the object's centre of mass are related by 
\begin{subequations}
	\label{eq:coupledkinematics}
	\begin{align}
		p_{i}(s) &= p_{\scriptscriptstyle O}(s) + p_{i \scriptscriptstyle O}(s) = p_{\scriptscriptstyle O}(s) + R_{{\scriptscriptstyle E_i}}(s) p_{i \scriptscriptstyle O}^{{\scriptscriptstyle E_i}}, \label{eq:kin_1}\\
		\phi_{i}(s) &= \phi_{\scriptscriptstyle O}(s) + \phi_{i \scriptscriptstyle O},
	\end{align}
\end{subequations}
where $p_{i \scriptscriptstyle O}^{{\scriptscriptstyle E_i}}$ and $\phi_{i \scriptscriptstyle O}$ are the constant distance and orientation offsets between the reference frames $\{O\}$ and $\{E_i\}$, respectively, and are considered to be known.
The $3\times 3$ matrix $R_{{\scriptscriptstyle E_i}}$ denotes the rotation matrix of the reference frame $\{E_i\}$ with respect to $\{I\}$. 

\subsection{Manipulator Dynamics}
The equations of motion for the $i$th manipulator with $n_i$ DOF and the joint variables $q_{i} \in \mathbb{R}^{n_i}$ can be described by \citep{siciliano2010robotics}
\begin{equation} \label{eq:dynamics_manipulator}
	M_{i}(q_{i})\ddot{q}_{i}+C_{i}(q_{i},\dot{q}_{i})\dot{q}_{i}+g_{i}(q_{i})=\tau_{i} - J_{i}(q_i)^T h_{i},
\end{equation} 
where $\tau_{i} \in \mathbb{R}^{n_i}$ are the joint torques, $M_{i}(q_{i}) \in \mathbb{R}^{n_i \times n_i}$ is the positive definite inertia matrix, $C_{i}(q_{i},\dot{q}_{i}) \dot{q}_{i} \in \mathbb{R}^{n_i \times 1}$ is the vector of Coriolis and centrifugal forces, $g_{i}(q_{i}) \in \mathbb{R}^{n_i}$ is the vector of gravitational terms, $J_{i} \in \mathbb{R}^{6 \times n_i}$ is the $i$th manipulator's geometric Jacobian, and $h_{i} \in \mathbb{R}^6$ is the vector of generalized forces exerted by the $i$th end-effector on the object.
A vector of generalized forces acting on a rigid body consists of a linear component (pure force) and an angular component (pure moment). We will refer to a force/moment pair as a \textit{wrench}.
In \eqref{eq:dynamics_manipulator}, the torques $J_{i}^{T}(q_i) h_{i}$ are a portion of the actuation torques that is needed to balance the torques induced at the joints by the contact forces.


\subsection{Object Dynamics}
Regarding the object, the pose and the velocity of its centre of mass are denoted by $x_{\scriptscriptstyle O} = [p_{\scriptscriptstyle O}^T, \phi_{\scriptscriptstyle O}^T]^T$ and $v_{\scriptscriptstyle O} = [\dot{p}_{\scriptscriptstyle O}^T, \omega_{\scriptscriptstyle O}^T]^T$, respectively, where $\omega_{\scriptscriptstyle O} \in \mathbb{R}^{3}$ is the object's angular velocity.
The following second-order dynamics, which is based on the Newton-Euler formulation can be considered for the object:
\begin{equation} \label{eq:dynamics_object}
	M_{\scriptscriptstyle O}(x_{\scriptscriptstyle O})\dot{v}_{\scriptscriptstyle O} + C_{\scriptscriptstyle O}(x_{\scriptscriptstyle O},\dot{x}_{\scriptscriptstyle O})v_{\scriptscriptstyle O} + g_{\scriptscriptstyle O} = h_{\scriptscriptstyle O},
\end{equation}
where 
\begin{subequations}
	\begin{align}
		M_{\scriptscriptstyle O} &= \begin{bmatrix}
			m I_3 & 0_{3 \times 3} \\[1pt] 
			0_{3 \times 3} & I_{\scriptscriptstyle O}
		\end{bmatrix}, \\
		C_{\scriptscriptstyle O} &= \begin{bmatrix}
			0_{3 \times 3} & 0_{3 \times 3} \\[1pt] 
			0_{3 \times 3} & S(\omega_{\scriptscriptstyle O}) I_{\scriptscriptstyle O}
		\end{bmatrix}, \\
		g_{\scriptscriptstyle O} &= \begin{bmatrix}
			- m g \\ 
			0_3
		\end{bmatrix},
	\end{align}
\end{subequations}
and where $m$ is the mass of the object, $g \in \mathbb{R}^{3}$ is the gravitational acceleration vector, $I_{\scriptscriptstyle O} \in \mathbb{R}^{3 \times 3}$ is the object's inertia tensor relative to the centre of mass when expressed in a frame parallel to $\{I\}$ with origin at the centre of mass, and $h_{\scriptscriptstyle O} \in \mathbb{R}^6$ is the wrench acting on the object's centre of mass. Here, $S(\cdot)$ is the skew-symmetric matrix operator performing the cross product, i.e., $S(u)w = u \times w$ for $u,w \in \mathbb{R}^3$.


\section{Time-optimal cooperative path tacking with rigid contacts}\label{sec:mainResult}
In this section, we show that the time-optimal path tracking problem for cooperative manipulators rigidly grasping an object can be cast as a convex optimization problem. 
The first step is to rewrite the manipulator dynamics in terms of $\ddot{s}, \dot{s}^2$ and $s$.

Given $p_{\scriptscriptstyle O}(s)$ and $\phi_{\scriptscriptstyle O}(s)$, the pose of each end-effector can be computed by \eqref{eq:coupledkinematics}. Then, inverse kinematics can be employed to compute the corresponding joint variables as functions of the path coordinate $s$. Using the chain rule, the joint velocities and accelerations of the $i$th manipulator can be written as
\begin{subequations}\label{eq_chain}
	\begin{align} 
		\dot{q}_{i}(s) &= {q}'_{i}(s) \dot{s}, \label{eq_qdot} \\
		\ddot{q}_{i}(s) &= {q}'_{i}(s) \ddot{s} + {q}''_{i}(s) \dot{s}^2, \label{eq_qdoubledot}
	\end{align}
\end{subequations}
where ${q}'_{i}(s) = \partial q_{i}(s)/\partial s$ and ${q}''_{i}(s) = \partial^2q_{i}(s)/\partial s^2$. Substituting these expressions in \eqref{eq:dynamics_manipulator} yields the following expression for the equations of motion:
\begin{equation}\label{eq:general:tau_i}
	\tau_{i}(s) = m_{i}(s) \ddot{s} + c_{i}(s) \dot{s}^2 + g_{i}(s) + J_{i}(s)^T h_{i}(s),
\end{equation}
where
\begin{subequations}
	\begin{align}
		m_{i}(s) &= M_{i}(q_{i}(s)) {q}'_{i}(s), \\
		c_{i}(s) &= M_{i}(q_{i}(s)) {q}''_{i}(s) + C_{i}(q_{i}(s),{q}'_{i}(s)) {q}'_{i}(s),
	\end{align}
\end{subequations}
for each $i \in \mathcal{N}$.

The second step is to show that the object dynamics in \eqref{eq:dynamics_object}, similar to Equation \ref{eq:general:tau_i} can be written as a function of $\ddot{s}, \dot{s}^2$ and $s$. This allows us to use $\ddot{s}$ and $\dot{s}^2$ as optimization variables later. To this end, we start by writing the first- and the second-order time derivatives of the pose of the object's centre of mass as
\begin{subequations}\label{eq:x_o}
	\begin{align}
		\dot{x}_{\scriptscriptstyle O}(s) &= {x}_{\scriptscriptstyle O}'(s) \dot{s}, \\
		\ddot{x}_{\scriptscriptstyle O}(s) &= {x}_{\scriptscriptstyle O}'(s) \ddot{s} + {x}_{\scriptscriptstyle O}''(s) \dot{s}^2, 
	\end{align} 
\end{subequations}
respectively, where ${x}_{\scriptscriptstyle O}'(s) = \partial x_{\scriptscriptstyle O}(s)/\partial s$ and ${x}_{\scriptscriptstyle O}''(s) = \partial^2x_{\scriptscriptstyle O}(s)/\partial s^2$.
Next, we find the relationship between the velocity and the time derivative of the pose of the object's centre of mass. For this, we know that it is possible to find the relationship between the object's angular velocity $\omega_{\scriptscriptstyle O}$ and its rotational velocity $\dot{\phi}_{\scriptscriptstyle O}$ for a given set of orientation angles \citep{siciliano2010robotics}. Let $T(\phi_{\scriptscriptstyle O})$ be the transformation between these two velocities, i.e., $\omega_{\scriptscriptstyle O} = T(\phi_{\scriptscriptstyle O}) \dot{\phi}_{\scriptscriptstyle O}$. Using this equation, the velocity of the object's centre of mass can be written as 
\begin{equation}\label{eq:v_o}
	v_{\scriptscriptstyle O}(s) = T_{\scriptscriptstyle O}(s) \dot{x}_{\scriptscriptstyle O}(s),
\end{equation}
where
\begin{equation}
	T_{\scriptscriptstyle O}(s) = \begin{bmatrix}
		I_3 & 0_{3 \times 3} \\[1pt]  
		0_{3 \times 3} &  T(\phi_{\scriptscriptstyle O}(s))
	\end{bmatrix}.
\end{equation}
Substituting \eqref{eq:x_o} into \eqref{eq:v_o} and its derivative, yields the following expressions for $v_{\scriptscriptstyle O}$ and $\dot{v}_{\scriptscriptstyle O}$, respectively:
\begin{subequations}\label{eq:v_o_dot}
	\begin{align}
		v_{\scriptscriptstyle O}(s) &= T_{\scriptscriptstyle O}(s) {x}_{\scriptscriptstyle O}'(s) \dot{s}, \\
		\dot{v}_{\scriptscriptstyle O}(s) &= \Big(\frac{\partial T_{\scriptscriptstyle O}(s)}{\partial s} x_{\scriptscriptstyle O}'(s)+T_{\scriptscriptstyle O}(s) x_{\scriptscriptstyle O}''(s)\Big)\dot{s}^2 \notag \\
		 & \quad + T_{\scriptscriptstyle O}(s) x_{\scriptscriptstyle O}'(s) \ddot{s}.
	\end{align} 
\end{subequations}
Finally, substituting \eqref{eq:x_o} and \eqref{eq:v_o_dot} into \eqref{eq:dynamics_object}, an expression for the object dynamics can be obtained as follows:
\begin{equation}\label{eq:object_ho}
	m_{\scriptscriptstyle O}(s)\ddot{s} + c_{\scriptscriptstyle O}(s)\dot{s}^2 + g_{\scriptscriptstyle O} = h_{\scriptscriptstyle O}(s),
\end{equation}
where 
\begin{subequations}
	\begin{align}
		m_{\scriptscriptstyle O}(s) &= M_{\scriptscriptstyle O}(x_{\scriptscriptstyle O}(s)) T_{\scriptscriptstyle O}(s) x_{\scriptscriptstyle O}'(s), \\
		c_{\scriptscriptstyle O}(s) &= M_{\scriptscriptstyle O}(x_{\scriptscriptstyle O}(s)) \frac{\partial T_{\scriptscriptstyle O}(s)}{\partial s} x_{\scriptscriptstyle O}'(s) \notag \\
		& \quad + M_{\scriptscriptstyle O}(x_{\scriptscriptstyle O}(s)) T_{\scriptscriptstyle O}(s) x_{\scriptscriptstyle O}''(s) \\  & \quad +  C_{\scriptscriptstyle O}(x_{\scriptscriptstyle O}(s),x_{\scriptscriptstyle O}'(s)) T_{\scriptscriptstyle O}(s) x_{\scriptscriptstyle O}'(s). \notag
	\end{align}
\end{subequations}

The final step concerns the relationship between the wrench $h_{\scriptscriptstyle O}$ acting on the object's centre of mass and the wrenches $h_i, i \in \mathcal{N}$ exerted by the end-effectors to the object at the grasping points. 
Let $v_{i} = [\dot{p}_{i}^T, \omega_{i}^T]^T$ be the velocity of the $i$th end-effector, where $\omega_{i} \in \mathbb{R}^{3}$ is the angular velocity. Differentiation of \eqref{eq:kin_1} together with the fact that $\omega_i = \omega_{\scriptscriptstyle O}$ due to the grasping rigidity, leads to 
\begin{equation}\label{eq:vi_vo}
	v_i(s) = J_{{\scriptscriptstyle O}_i}(s) v_{\scriptscriptstyle O}(s), \quad \forall i \in \mathcal{N}, 
\end{equation}
where $J_{{\scriptscriptstyle O}_i}$ is called the object-to-agent Jacobian matrix, with
\begin{equation}
	J_{{\scriptscriptstyle O}_i}(s) = \begin{bmatrix}
		I_3	& -S(p_{i \scriptscriptstyle O}(s)) \\[1pt]  
		0_{3 \times 3}	& I_3
	\end{bmatrix},
\end{equation}
$\forall i \in \mathcal{N}$, which is full-rank due to the grasp rigidity. The \textit{grasp matrix} $G \in \mathbb{R}^{6 \times 6N}$ is formed by stacking $J_{{\scriptscriptstyle O}_i}(s)^T, i \in \mathcal{N}$ as
\begin{equation}
	G(s) = [J_{{\scriptscriptstyle O}_1}(s)^T \cdots J_{{\scriptscriptstyle O}_N}(s)^T],
\end{equation}
and has full row rank. 
Let $v(s) = [v_1(s)^T \cdots v_N(s)^T]^T$.
Equation \eqref{eq:vi_vo} can now be written in the following form:
\begin{equation}
	v(s) = G(s)^T v_{\scriptscriptstyle O}(s).
\end{equation}
Let $h = [h_1^T \cdots h_N^T]^T$. The kineto-statics duality \citep{siciliano2010robotics} together with the grasp rigidity suggest that the wrenches $h_{\scriptscriptstyle O}$ and $h_i, i \in \mathcal{N}$ are related by
\begin{align}\label{eq:h_O}
		h_{\scriptscriptstyle O}(s) &= G(s) h(s)  \notag \\
		&= \sum_{i \in \mathcal{N}}^{} G_i(s) h_i(s),
\end{align}
where $G_i$ is the $i$th block of the grasp matrix, i.e., $G_i = J_{{\scriptscriptstyle O}_i}^T$.

Finally, having the dynamic equations of the manipulators and the object written in terms of $\ddot{s}, \dot{s}^2$ and $s$, and having the relationship between the wrenches $h_{\scriptscriptstyle O}$ and $h_i, i \in \mathcal{N}$, we are ready to present the convex formulation of the problem of interest.
From the equations in \eqref{eq:general:tau_i}, \eqref{eq:object_ho} and \eqref{eq:h_O} it can be seen that choosing $a(s) = \ddot{s}$ and $b(s) = \dot{s}^2$ together with $h_{\scriptscriptstyle O}(s)$ and $h_i(s)$, $\tau_{i}(s), i \in \mathcal{N}$ as optimization variables, makes these equations, which enter as constraints in the optimization problem, linear in the optimization variables. 
The objective function for our problem of interest is the same as \eqref{op:obj}, which is convex in $b(s)$. With the introduced optimization variables, the problem of interest can be cast as the following convex optimization problem:
\begin{mini*}[2]
	{
	}{\int_{0}^{1}\frac{1}{\sqrt{b(s)}}ds}
	{}{}
	\addConstraint{\tau_{i}(s) = 
		m_{i}(s)  a(s) + c_{i}(s) b(s) + g_{i}(s) }{} \notag
	\addConstraint{\qquad \quad + J_{i}(s)^T h_{i}(s),}{\quad i \in \mathcal{N},}
	\addConstraint{h_{\scriptscriptstyle O}(s) = m_{\scriptscriptstyle O}(s)a(s) + c_{\scriptscriptstyle O}(s)b(s) + g_{\scriptscriptstyle O},}{} 
	\addConstraint{h_{\scriptscriptstyle O}(s) = \sum_{i \in \mathcal{N}}^{} G_i(s) h_i(s),}{}
	\addConstraint{b(0)}{=\dot{s}_0^2, \quad b(1)=\dot{s}_T^2,}
	\labelOP{op:general:convex}
	\addConstraint{0 \leq b(s) \leq \overline{b}(s),}{}
	\addConstraint{{b}'(s)}{=2a(s),}
	\addConstraint{\underline{\tau}_i(s) \leq \tau_i(s) \leq \overline{\tau}_i(s),}{\quad i \in \mathcal{N},}
	\addConstraint{\forall s \in [0,1],}{}
\end{mini*}
with variables $a, b, h_{\scriptscriptstyle O}$ and $\tau_i, h_i, i \in \mathcal{N}$.
In \eqref{op:general:convex}, $\underline{\tau}_i$ and $\overline{\tau}_i$ are the lower and upper bounds on the $i$th manipulator's joint torques, respectively. This problem can finally be reformulated as an SOCP using the direct transcription method and a procedure similar to the one used in \citet{verscheure2009time}.

Note that to consider all possible combinations of wrenches when finding the optimal solution, the relationship between the wrenches $h_{\scriptscriptstyle O}$ and $h_i, i \in \mathcal{N}$ is intentionally kept as \eqref{eq:h_O}, instead of solving \eqref{eq:h_O} for $h_i, i \in \mathcal{N}$ and representing $h_i, i \in \mathcal{N}$ based on a particular wrench distribution.
For a cooperative manipulation task, in general, there exists an infinite number of wrenches exerted by the end-effectors that result in the same desired wrench on the centre of mass of the object. This is due to multiple manipulators being involved in the manipulation task. 
When using a predefined wrench distribution, this redundancy is lost.
Our approach uses the full actuation available to the system and optimally handles the actuation redundancy.
Compared to a case where a particular wrench distribution is used, a smaller minimal traversal time is expected from our approach.
A numerical comparison using some common wrench distributions is carried out and reported in Section \ref{sec:simulation} which confirms this point.



\section{Extension to grasping with friction}\label{sec:friction}
In this section we relax the assumption on the rigid grasping points. 
We consider the problem stated in Section \ref{sec:probleStatement}, but with a different type of grasp.
In particular, we assume that contacts between the manipulators and the object are of \textit{point contact with friction} or \textit{soft-finger} types.
Also, we assume a \textit{force-closure} grasp, which will be defined later.
For grasps using frictional forces, a model for friction must be provided. For this purpose, we will use a simple model which is referred to as the \textit{Coulomb friction model}. This model is an empirical model which states that the allowed force in the tangential directions to a surface is proportional to the applied force at the normal direction. The constant of proportionality which is referred to as the coefficient of friction, is determined by the materials that are in contact.
A point contact with friction model is used when friction exists between the manipulator tip and the object, in which case forces can be exerted in
any direction that is within the friction cone for the contact.
For a soft-finger contact, not only forces are allowed to be applied in a cone about the surface normal, but also torques about that normal \citep{murray2017mathematical}.
These types of contacts complicate the problem since each contact may only apply a wrench that respects friction cone constraints to prevent slip, instead of an arbitrary wrench associated with rigid contacts. 

\begin{figure}[!t]
	\centering
	\includegraphics[width=3.1in]{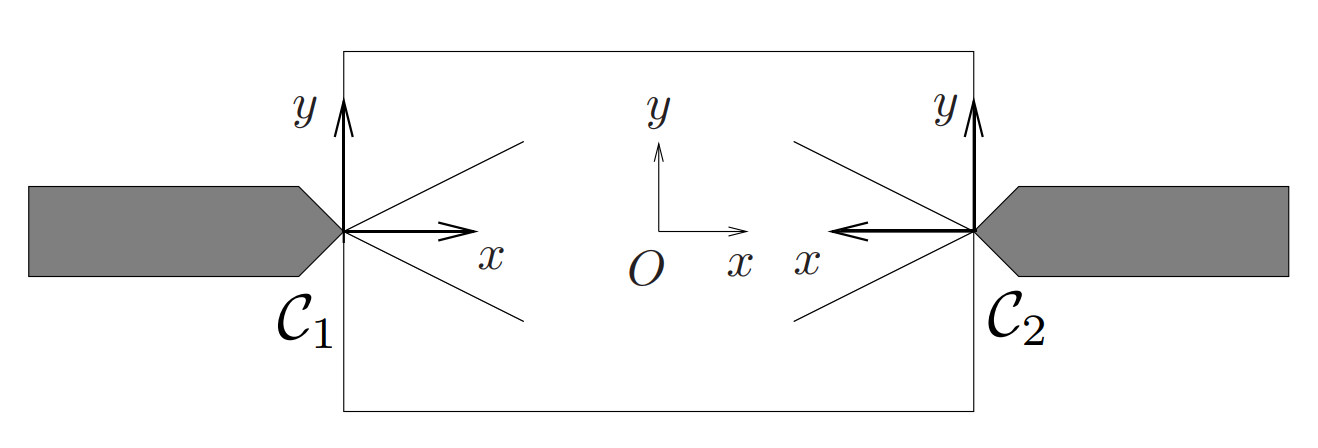}
	\caption{Friction cones for a planar grasping.}
	\label{fig:FCplanar}
\end{figure}

We start by describing the grasp.
The $i$th contact can be modeled using a wrench basis, $B_{\scriptscriptstyle C_i} \in \mathbb{R}^{6\times m_i}$, 
and a friction cone, $FC_{\scriptscriptstyle C_i}$ \citep{murray2017mathematical}. 
The dimension of the wrench basis, $m_i$, indicates the
number of independent forces/moments that can be applied by the contact.
Let us consider the planar grasp shown in \figurename~\ref{fig:FCplanar}, with point contacts with friction.
For convenience, the contact coordinate frame ${\mathcal C}_i$, is chosen in such a way that its $x$-axis points in the direction of the inward surface normal at the point of contact. 
Let $h_{\scriptscriptstyle C_i}$ be the wrench applied at the $i$th contact point.
We represent this wrench with respect to a basis of directions which are consistent with the friction model:
\begin{align}\label{eq:B_ci}
	h_{\scriptscriptstyle C_i} &= B_{\scriptscriptstyle C_i} f_{\scriptscriptstyle C_i} \notag \\ 
	&= \begin{bmatrix}
		1 & 0 & 0 & 0 & 0 & 0 \\
		0 & 1 & 0 & 0 & 0 & 0 \\
	\end{bmatrix}^T f_{\scriptscriptstyle C_i}, \quad f_{\scriptscriptstyle C_i} \in FC_{\scriptscriptstyle C_i},
\end{align}
where the friction cone for the $i$th contact is
\begin{equation}\label{eq:FCplanar}
	FC_{\scriptscriptstyle C_i} = \left\{ f=\left [ f_x \ f_y \right ]^T \in \mathbb{R}^2 : \left| f_y \right| \leq \mu f_x, f_x \geq 0 \right\},
\end{equation}
In \eqref{eq:FCplanar}, $f_x$ and $f_y$ are the normal and tangential force components, respectively, and $\mu > 0$ is the coefficient of friction.

For a soft-finger contact with a coordinate frame where its $x$-axis points in the direction of the inward surface normal, the wrench basis and the friction cone are as follows:
\begin{subequations}
	\begin{align}
		&B_{\scriptscriptstyle C_i} = \begin{bmatrix}
			1 & 0 & 0 & 0 & 0 & 0 \\
			0 & 1 & 0 & 0 & 0 & 0 \\
			0 & 0 & 1 & 0 & 0 & 0 \\
			0 & 0 & 0 & 1 & 0 & 0 \\
		\end{bmatrix}^T, \\
	    &FC_{\scriptscriptstyle C_i} = \notag \\
	    & \ \  \left\{ f \in \mathbb{R}^4 : \sqrt{f_y^2+f_z^2} \leq \mu f_x, f_x \geq 0, \left| t_x \right| \leq \gamma f_x \right\}, \label{eq:FC_soft}
		\end{align}
\end{subequations}
where $f=\left [ f_x , f_y , f_z , t_x \right ]^T$, and where $\gamma > 0$ is the coefficient of torsional friction.

The \textit{contact map}, $\bar{G}_i \in \mathbb{R}^{6\times m_i}$, is defined to be the linear map between the $i$th contact wrench represented with respect to $B_{\scriptscriptstyle C_i}$, 
and the object wrench at the centre of mass:
\begin{equation}
	\bar{G}_i = G_i B_{\scriptscriptstyle C_i}.
\end{equation}
With this definition, the object wrench can be written as 
\begin{equation}
	h_{\scriptscriptstyle O} = \bar{G} f_{\scriptscriptstyle C},  \quad f_{\scriptscriptstyle C} \in FC,
\end{equation}
where 
\begin{equation}
	\bar{G} = \begin{bmatrix}
		G_1 B_{\scriptscriptstyle C_1}  \cdots   G_N B_{\scriptscriptstyle C_N} \\
	\end{bmatrix},
\end{equation}
is called \textit{grasp map}, and where
\begin{align}
	f_{\scriptscriptstyle C} &= \begin{bmatrix}
		f_{\scriptscriptstyle C_1}^T  \cdots   f_{\scriptscriptstyle C_N}^T \\
	\end{bmatrix}^T \in \mathbb{R}^m, \\
	FC &= FC_{\scriptscriptstyle C_1} \times \cdots \times FC_{\scriptscriptstyle C_N} \subset \mathbb{R}^m, \\
	m &= m_1 + \cdots + m_N. 
\end{align}
The full friction cone, $FC$, is the Cartesian product of all the friction cones.
A grasp is completely described by the grasp map $\bar{G}$ and the friction cone $FC$.

Wrenches in the null space of $\bar{G}$ correspond to those wrenches that can be exerted at the contact points without resulting in a net wrench on the object. These are referred to as \textit{internal forces}.
For collaborative manipulation with the mentioned types of contacts, it is critical to ensure that the object does not slip. To ensure this, each contact wrench must remain in the respective friction cone.
Internal forces can be used to insure that contact wrenches satisfy friction cone constraints. Any given vector of contact wrenches can be brought into the friction cone by adding a sufficiently large wrench in the null space of $\bar{G}$, if the condition $N(\bar{G}) \cap \text{int}(FC) \neq \varnothing$, where $N(\cdot)$ denotes the null space of a matrix and $\text{int}(FC)$ is the interior of the friction cone, is satisfied \citep{cole1988kinematics}.
In order to be able to firmly grasp an object, it is desirable that internal forces exist and lie in the interior of the friction cone.

It is also desirable that any given wrench on the object can be achieved by an appropriate choice of contact wrenches lying in the friction cone. The mathematical characterization of this ability is $\bar{G}(FC)=\mathbb{R}^6$, i.e., the grasp map $\bar{G}$ should map the friction cone $FC$ onto $\mathbb{R}^6$. This ability is linked to the ability of a grasp to resist any applied wrench, which is called force-closure. A grasp is a \textit{force-closure} grasp if given any external wrench $h_e \in \mathbb{R}^6$ applied to the object, there exist contact wrenches $f_{\scriptscriptstyle C} \in FC$ such that $\bar{G} f_{\scriptscriptstyle C} = - h_e$. It follows directly from this definition that a grasp is force-closure if and only if $\bar{G}(FC)=\mathbb{R}^6$ \citep{murray2017mathematical}.
It is also shown in \citet{murray2017mathematical} that the existence of an internal force belonging to the interior of the friction cone is a necessary condition for a grasp to be force-closure.

In order to have a grasp with the two desirable properties mentioned above, i.e., existence of an internal force which belongs to the interior of the friction cone and $\bar{G}(FC)=\mathbb{R}^6$, we assume that the grasp in our problem is a force-closure grasp. Constructing force-closure grasps is not the focus of this paper and the reader is referred to the related work on this subject, see e.g., \citet{nguyen1988constructing,ponce1995computing}. One particularly simple method for the case of two contact points is proposed in \citet{nguyen1988constructing}. A planar grasp with two point contacts with friction is force-closure if and only if the line connecting the contact points lies inside both friction cones. This result can be extended to the case of a spatial grasp with two soft-finger contacts.

Now, we are ready to present the optimization problem associated with the problem of interest in this section. 
In order to ensure that slipping does not occur, the friction cone constraints will enter as constraints in the optimization problem, and it can be seen from \eqref{eq:FCplanar} and \eqref{eq:FC_soft} that these constraints are convex. This follows from the fact that every norm on $\mathbb{R}^{n}$ is convex \citep{boyd2004convex}. In order to consider internal forces, we divide the wrench applied at the $i$th contact point into two parts: a motion-inducing part, denoted by $h_{\scriptscriptstyle M_i}$, and a part associated with internal forces, denoted by $h_{\scriptscriptstyle I_i}$. In other words, $h_{\scriptscriptstyle C_i} = h_{\scriptscriptstyle M_i} + h_{\scriptscriptstyle I_i}, \ i \in \mathcal{N}$, where $h_{\scriptscriptstyle I} = [h_{\scriptscriptstyle I_1}^T \cdots h_{\scriptscriptstyle I_N}^T]^T \in N(G)$. 
The constraint on $h_{\scriptscriptstyle I}$ being in the null space of $G$ can be written as $\sum_{i \in \mathcal{N}}^{} G_i h_{\scriptscriptstyle I_i} = 0$.
Let $f_{\scriptscriptstyle M_i}$ and $f_{\scriptscriptstyle I_i}$ be the contact wrenches corresponding to $h_{\scriptscriptstyle M_i}$ and $h_{\scriptscriptstyle I_i}$, represented with respect to $B_{\scriptscriptstyle C_i}$, respectively, i.e., $h_{\scriptscriptstyle M_i} = B_{\scriptscriptstyle C_i} f_{\scriptscriptstyle M_i}$ and $h_{\scriptscriptstyle I_i} = B_{\scriptscriptstyle C_i} f_{\scriptscriptstyle I_i}$. As mentioned above, in order to have a firm grasp, it is desirable that the internal forces lie in the interior of the friction cone. The constraints $f_{\scriptscriptstyle I_i} \in \text{int}({FC}_{\scriptscriptstyle C_i}), \ i \in \mathcal{N}$ will be added to the optimization problem for this purpose.
Note that to prevent slip, $f_{\scriptscriptstyle C_i} = f_{\scriptscriptstyle M_i} + f_{\scriptscriptstyle I_i}$ must also respect the friction cone constraints.
By introducing $h_{\scriptscriptstyle M_i}, h_{\scriptscriptstyle I_i}, f_{\scriptscriptstyle M_i}$ and $f_{\scriptscriptstyle I_i}$ as additional optimization variables and with the discussed changes, the problem of interest in this section can be cast as the following convex optimization problem:
\begin{mini}[2]<b>
	{
	}{\int_{0}^{1}\frac{1}{\sqrt{b(s)}}ds}
	{}{}
	\addConstraint{\tau_{i}(s) = 
		m_{i}(s)  a(s) + c_{i}(s) b(s) + g_{i}(s) }{} \notag
	\addConstraint{\qquad \quad + J_{i}(s)^T (h_{\scriptscriptstyle M_i}(s) + h_{\scriptscriptstyle I_i}(s)),}{} \notag
	\addConstraint{\qquad \qquad i \in \mathcal{N},}{} \notag
	\addConstraint{h_{\scriptscriptstyle O}(s) = m_{\scriptscriptstyle O}(s)a(s) + c_{\scriptscriptstyle O}(s)b(s) + g_{\scriptscriptstyle O},}{} 
	\addConstraint{h_{\scriptscriptstyle O}(s) = \sum_{i \in \mathcal{N}}^{} G_i(s) (h_{\scriptscriptstyle M_i}(s) + h_{\scriptscriptstyle I_i}(s)),}{}
	\addConstraint{\sum_{i \in \mathcal{N}}^{} G_i(s) h_{\scriptscriptstyle I_i}(s)}{= 0,}
	\labelOP{op:FC:convex}
	\addConstraint{h_{\scriptscriptstyle M_i} = B_{\scriptscriptstyle C_i} f_{\scriptscriptstyle M_i},}{ \quad i \in \mathcal{N},}
	\addConstraint{h_{\scriptscriptstyle I_i} = B_{\scriptscriptstyle C_i} f_{\scriptscriptstyle I_i}}{\quad i \in \mathcal{N},}
	\addConstraint{(f_{\scriptscriptstyle M_i}(s) + f_{\scriptscriptstyle I_i}(s)) \in {FC}_{\scriptscriptstyle C_i}, \quad i \in \mathcal{N},}{}
	\addConstraint{f_{\scriptscriptstyle I_i}(s) \in \text{int}({FC}_{\scriptscriptstyle C_i}), \quad i \in \mathcal{N},}{}
	\addConstraint{b(0)}{=\dot{s}_0^2, \quad b(1)=\dot{s}_T^2,}
	\addConstraint{0 \leq b(s) \leq \overline{b}(s),}{}
	\addConstraint{{b}'(s)}{=2a(s),}
	\addConstraint{\underline{\tau}_i(s) \leq \tau_i(s) \leq \overline{\tau}_i(s),}{\quad i \in \mathcal{N},}
	\addConstraint{\forall s \in [0,1],}{}
\end{mini}
with variables $a, b, h_{\scriptscriptstyle O}$ and $\tau_i, h_{\scriptscriptstyle M_i}, h_{\scriptscriptstyle I_i}, f_{\scriptscriptstyle M_i}, f_{\scriptscriptstyle I_i}, i \in \mathcal{N}$.
This optimization problem is convex since the objective function and all the constraints are either linear or convex in the optimization variables. 
Again, this problem can be reformulated as an SOCP after employing the direct transcription method.



\section{Numerical Simulation}\label{sec:simulation}
In this section, we provide some numerical simulations to illustrate the results. We start by presenting the simulation results for the case of rigid contacts.
\subsection{Rigid Contacts}\label{sim:rigid}
Consider the problem setup described in Section \ref{sec:probleStatement} with two $6$ DOF Stanford manipulators (see \citet{siciliano2010robotics}, \figurename~2.25). Suppose the manipulators are located at $[0 \ -1.4 \ 0]^T$ and $[0 \ 1.4 \ 0]^T$ in $\{I\}$ and are placed in such a way that the rotation matrices of the base frames $\{B_i\}, i=1,2,$ with respect to $\{I\}$ are $R_{{\scriptscriptstyle B_1}} = I_3$ and $R_{{\scriptscriptstyle B_2}}= \text{diag}(-1,-1,1)$, where $\text{diag}$ is the diagonal matrix with the vector elements as diagonal elements. Also, suppose that the manipulators are rigidly grasping a cuboid object of dimensions $0.07 \times 0.07 \times 0.4$ m$^3$ and with mass $10$ kg. We assume that the centre of mass of the object is located at its centroid.
The dynamic parameters used for the simulation are given in Tables~\ref{tb:param_general_1}-\ref{tb:param_general_2}, where $m_{ij}$ and $l_{ij}$ denote the mass and the length of the $j$th link of the $i$th manipulator, respectively. 
We assume that all links except the third one are cylindrical with radius of 0.12, 0.1, 0.06, 0.05, 0.05 (all in meter), respectively, and compute the inertia tensors accordingly. The third link is assumed to be a cuboid with equal length and width of size 0.12 m. 
\begin{table}[b!]
	\caption{Masses of links in kg}\label{tb:param_general_1}
	\centering
	\begin{tabular}{|c|cccccc|} \hline
		& $m_{i1}$ & $m_{i2}$ & $m_{i3}$ & $m_{i4}$ & $m_{i5}$ & $m_{i6}$ \\ \hline 
		 $i=1,2$ & 15 & 10 & 8 & 1 & 0.7 & 0.5 \\ \hline
	\end{tabular}
\end{table}
\begin{table}[b!]
	\caption{Lengths of links in m}\label{tb:param_general_2}
	\centering
	\begin{tabular}{|c|ccccc|} \hline
		& $l_{i1}$ & $l_{i2}$  & $l_{i4}$ & $l_{i5}$ & $l_{i6}$ \\ \hline 
		$i=1,2$ & 0.6 & 0.5 & 0.15 & 0.12 & 0.1 \\ \hline
	\end{tabular}
\end{table}
The desired path for the centre of mass of the object is given by
\begin{equation}
	p_{\scriptscriptstyle O}(s) = \begin{bmatrix}
		-0.3+1.2 s \\ 
		\sin(s-0.4) \\ 
		0.4+0.6 s
	\end{bmatrix}.
\end{equation}
For the orientation of the object, we have used 
ZYZ angles
$\phi_{\scriptscriptstyle O}(s) =[-0.5 s+0.2 \ 0.5 \sin(s) \ -0.4]^T$.
We assume that gravity acts along the negative $z$-axis.
For the simulation, the path coordinate $s$ is discretized on $80$ grid points.
The optimization problem is implemented and solved in SOCP form in MATLAB using the optimization modeling toolbox YALMIP \citep{lofberg2004yalmip}, and the solver MOSEK \citep{mosek}. The simulation is performed on a laptop with a 2.6 GHz Intel Core i7-5600U processor.

In the approach presented in this paper we need to compute the partial derivatives ${q}'_{i}(s) $ and ${q}''_{i}(s)$ at the descretization points. Any method that can provide these partial derivatives can be used for this purpose. For a manipulator with 6 DOF, computing these derivatives analytically can be computationally expensive. In this simulation, we are using a numerical approach to compute ${q}'_{i}(s) $ and ${q}''_{i}(s)$ at the discretization points. 

\begin{figure}[!t]
	\centering
	\includegraphics[width=11cm]{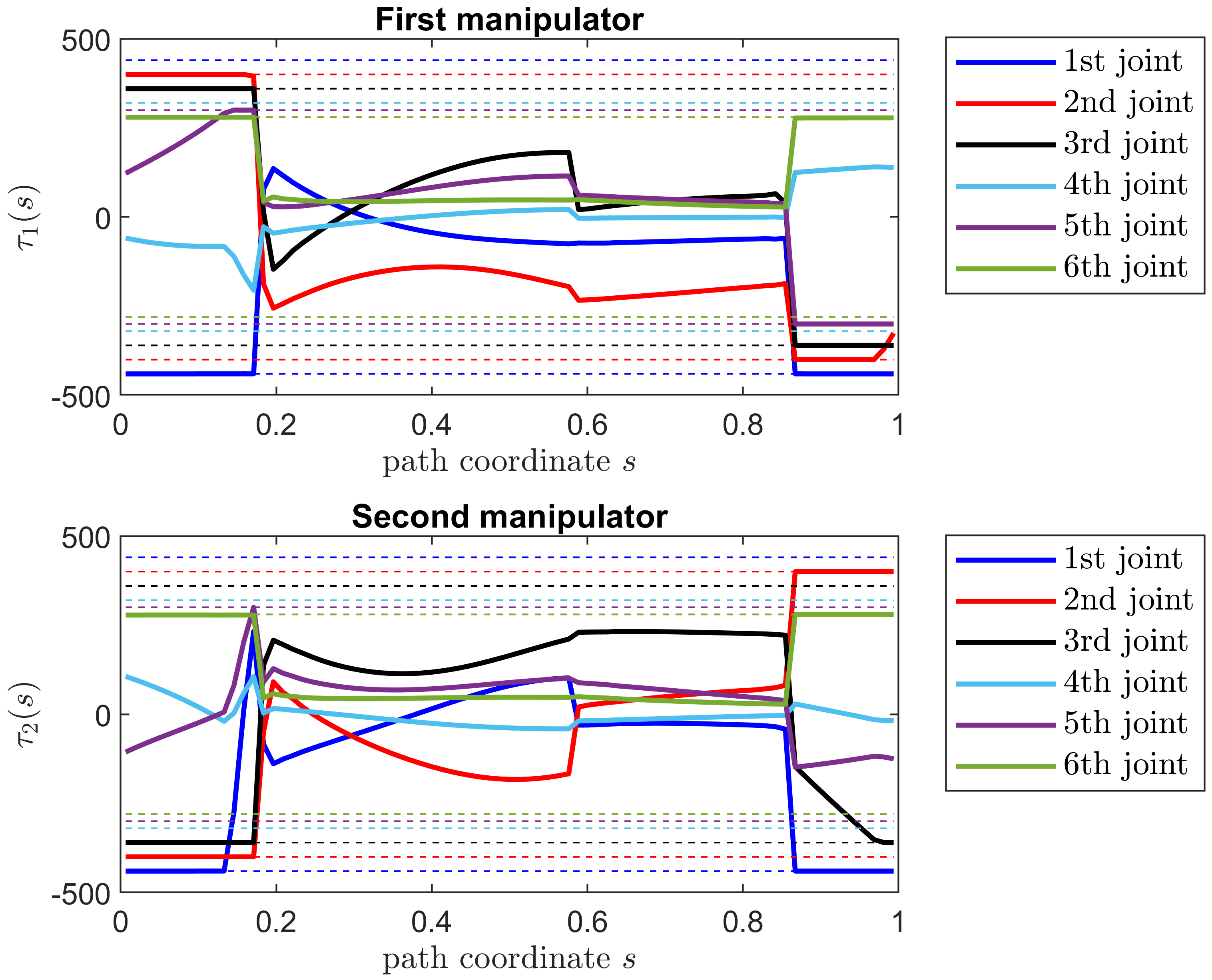}
	\caption{Joint torques/forces $\tau_{1}(s)$ and $\tau_{2}(s)$ for the case with rigid contacts. The units are Nm for a revolute and N for a prismatic joint.}
	\label{fig:torque_gen}
\end{figure}

\begin{figure}[!t]
	\centering
	\includegraphics[width=11cm]{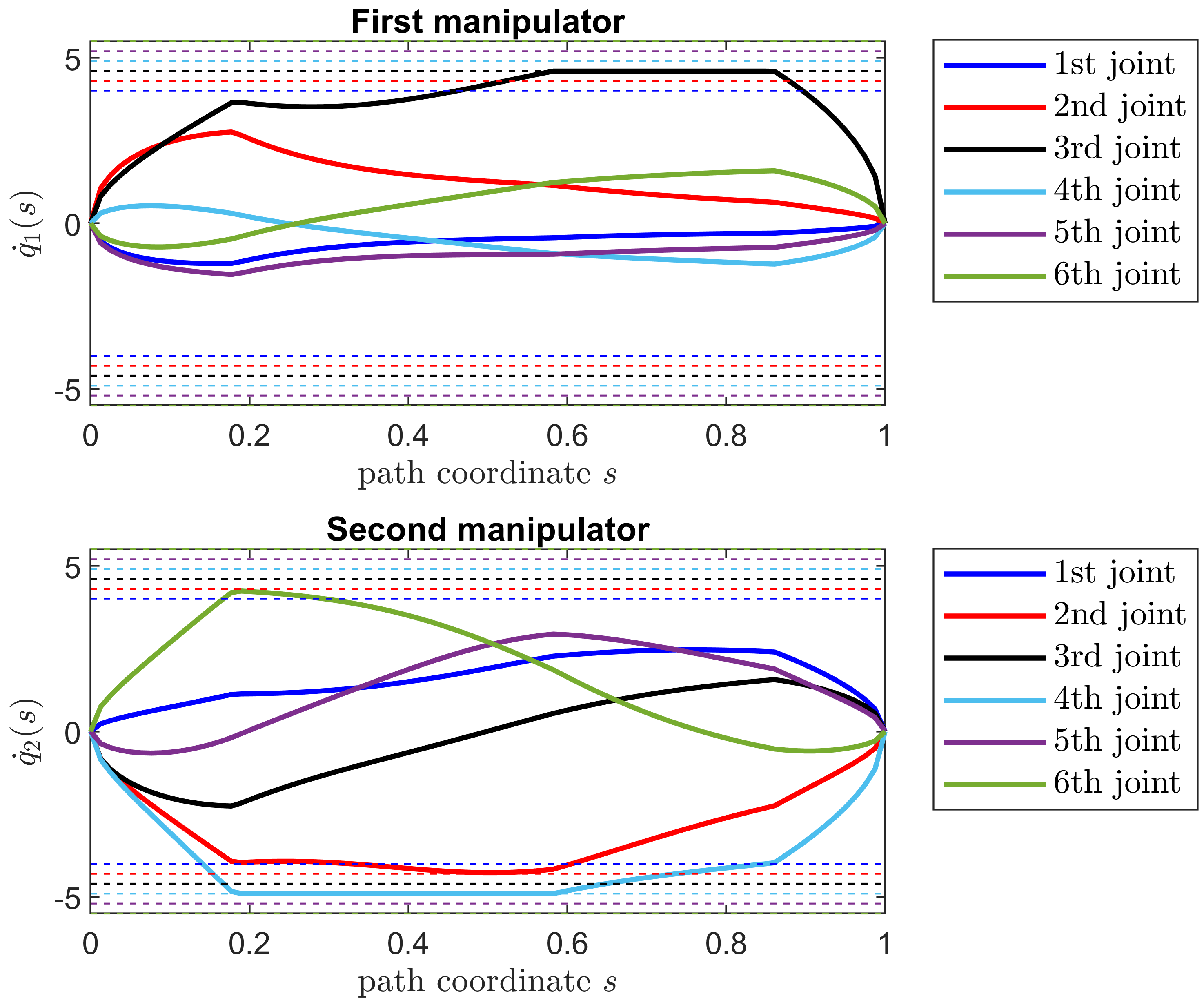}
	\caption{Joint velocities $\dot{q}_{1}(s)$ and $\dot{q}_{2}(s)$ for the case with rigid contacts. The units are rad/s for a revolute and m/s for a prismatic joint.}
	\label{fig:velocity_gen}
\end{figure}

The approach presented in this paper results in a minimal traversal time of $0.422$ sec for this simulation.
\figurename~\ref{fig:torque_gen} shows the resulting joint torques or forces and the corresponding bounds for both manipulators as functions of the path coordinate $s$. The bounds are depicted in the same color as the joint torque or force, but with a dashed line. Once the optimization variables $a$ and $b$ are obtained, Equations \eqref{eq_qdot} and \eqref{eq_qdoubledot} can be used to compute the joint velocities and accelerations, respectively. \figurename~\ref{fig:velocity_gen} shows the evolution of the joint velocities with respect to the path coordinate $s$ for both manipulators, where the initial and the final velocities have been set to zero. The considered bounds on the joint velocities can be seen in the same figure and are depicted with dashed lines in the same color as each joint velocity.

From the plots it can be seen that for every point along the path, either one of the joint torques/forces of the first or second manipulator is in saturation, or one of the joint velocities has reached its bound. 
This is in line with the findings in the work by \citet{chen1990structure}, where it is shown that the minimum-time control for multiple manipulators requires that at least one of the actuators is always saturated on any finite time subinterval, while the rest of them adjust their torques so that other constraints on the motion are not violated.

We have also performed the simulation for different grid sizes to get an idea of computation time. Solver time and YALMIP time are reported in Table~\ref{tb:solverTime} for grid sizes of $K = 30, 100, 300, 1000$. It can be seen that the increase in YALMIP time is greater than the increase in solver time.
\begin{table}[b!]
	\caption{Solver and YALMIP times in second for different grid sizes.}\label{tb:solverTime}
	\centering
	\begin{tabular}{|c|c|c|c|c|} \hline
		& K=30 & K=100 & K=300 &  K=1000  \\ \hline 
		Solver time & 0.573 & 0.600 & 0.701 & 0.982  \\ \hline
		YALMIP time & 0.280 & 0.425 & 2.191 & 3.355  \\ \hline
	\end{tabular}
\end{table}

Since all possible combinations of wrenches are considered in our approach, a smaller minimal traversal time from our approach is expected compared to a case where a particular wrench distribution is used. For the purpose of comparison, we have carried out some simulations with the same setup described above but different path and orientation for the object and with wrench distributions borrowed from other works. The resulting minimal traversal times using the wrench distribution from \citet{moon1997time}, \citet{verginis2022cooperative} and \citet[][Ch.29]{siciliano2008springer}, denoted by Ref. 1, Ref. 2 and Ref. 3, respectively, together with the results for the rigid and the frictional (see next subsection) contacts are reported in Table~\ref{tb:comparison}. 
In the simulations, the path coordinate $s$ is discretized on $50$ grid points.
The paths and the orientations that are used for this comparison can be seen in Tables~\ref{tb:pathPosition}-\ref{tb:pathOrientation}, where $p_{\scriptscriptstyle O}^x(s)$, $p_{\scriptscriptstyle O}^y(s)$ and $p_{\scriptscriptstyle O}^z(s)$ are the $x$, $y$ and $z$ components of the position of the object, respectively, and $[\varphi_{\scriptscriptstyle O}(s), \vartheta_{\scriptscriptstyle O}(s), \psi_{\scriptscriptstyle O}(s)]^T$ is the object's orientation in terms of ZYZ Euler angles. 
The considered bounds on joint torques and velocities are same as the ones in \figurename~\ref{fig:torque_gen} and \figurename~\ref{fig:velocity_gen}. 
In the tables, P.$i$ refers to the $i$th path.
As expected, the obtained minimal traversal time using our approach in the case of rigid contacts is significantly smaller than other values where a particular wrench distributions is used. This is because our approach exploits the full actuation available to the system.
%
%
%
%
%
%
\begin{table}[b!]
	\caption{Resulting minimal traversal times using the wrench distribution from \citet{moon1997time} (Ref.~1), \citet{verginis2022cooperative} (Ref.~2) and \citet[][Ch.29]{siciliano2008springer} (Ref.~3) together with the results from the rigid and the frictional cases, where the results are reported in seconds.}\label{tb:comparison}
	\centering
	\begin{tabular}{|c|c|c|c|c|c|} \hline
		& Rigid & Frictional & Ref. 1 & Ref. 2 & Ref. 3   \\ \hline 
		
		P.1 & 0.422 & 0.567 & 0.531 & 0.524 & 0.531 \\ \hline
		P.2 & 0.270 & 0.461 & 0.386 & 0.396 & 0.387  \\ \hline
		P.3 & 0.309 & 0.523 & 0.460 & 0.489 & 0.462 \\ \hline
		P.4 & 1.541 & 1.676 & 2.287 & 2.312 & 2.287 \\ \hline
		P.5 & 0.433 & 0.541 & 0.553 & 0.601 & 0.553 \\ \hline
		
	\end{tabular}
\end{table}
\begin{table}[b!]
	\caption{Paths used for the comparison.}\label{tb:pathPosition}
	\centering
	\begin{tabular}{|c|ccc|c|} \hline
		& $p_{\scriptscriptstyle O}^x(s)$ & $p_{\scriptscriptstyle O}^y(s)$ &  $p_{\scriptscriptstyle O}^z(s)$  \\ \hline 
		
		P.1 & $-0.3+1.2 s$  & $\sin(s-0.4)$ & $0.4+0.6 s$ \\ \hline
		P.2 & $\sin(s)$ & $\cos(s)-0.5$ & $0.3+0.7 s$  \\ \hline
		P.3 & $s^2$ & $\cos(s)-0.5$ & $0.7+s(s-1)$  \\ \hline
		P.4 & $s^2-0.3$ & $0.5 s^3$ & $1.2 (1-s)+0.4 $  \\ \hline
		P.5 & $-0.5+s$ & $0.4 s^3$ & $1.2 (1-s)+0.4$   \\ \hline
		
	\end{tabular}
\end{table}
\begin{table}[b!]
	\caption{Orientations used for the comparison.}\label{tb:pathOrientation}
	\centering
	\begin{tabular}{|c|ccc|c|} \hline
		& $\varphi_{\scriptscriptstyle O}(s)$ & $\vartheta_{\scriptscriptstyle O}(s)$ & $\psi_{\scriptscriptstyle O}(s)$  \\ \hline 
		
		P.1 & $-0.5 s+0.2$ & $0.5 \sin(s)$ & $-0.4$  \\ \hline
		P.2 & $0.5 s$ & $-0.5 s$ & $0$  \\ \hline
		P.3 & $0.5 s$ & $0.3 s$ & $-0.2 s$  \\ \hline
		P.4 & $0.5 \sin(s)$ & $-0.3 s$ & $0.3 s$  \\ \hline
		P.5 & $0.5 \cos(s)$ & $0.5$ & $0.3 s$  \\ \hline
		
	\end{tabular}
\end{table}

\subsection{Contacts with Friction}
\begin{figure}[!t]
	\centering
	\includegraphics[width=2.9in]{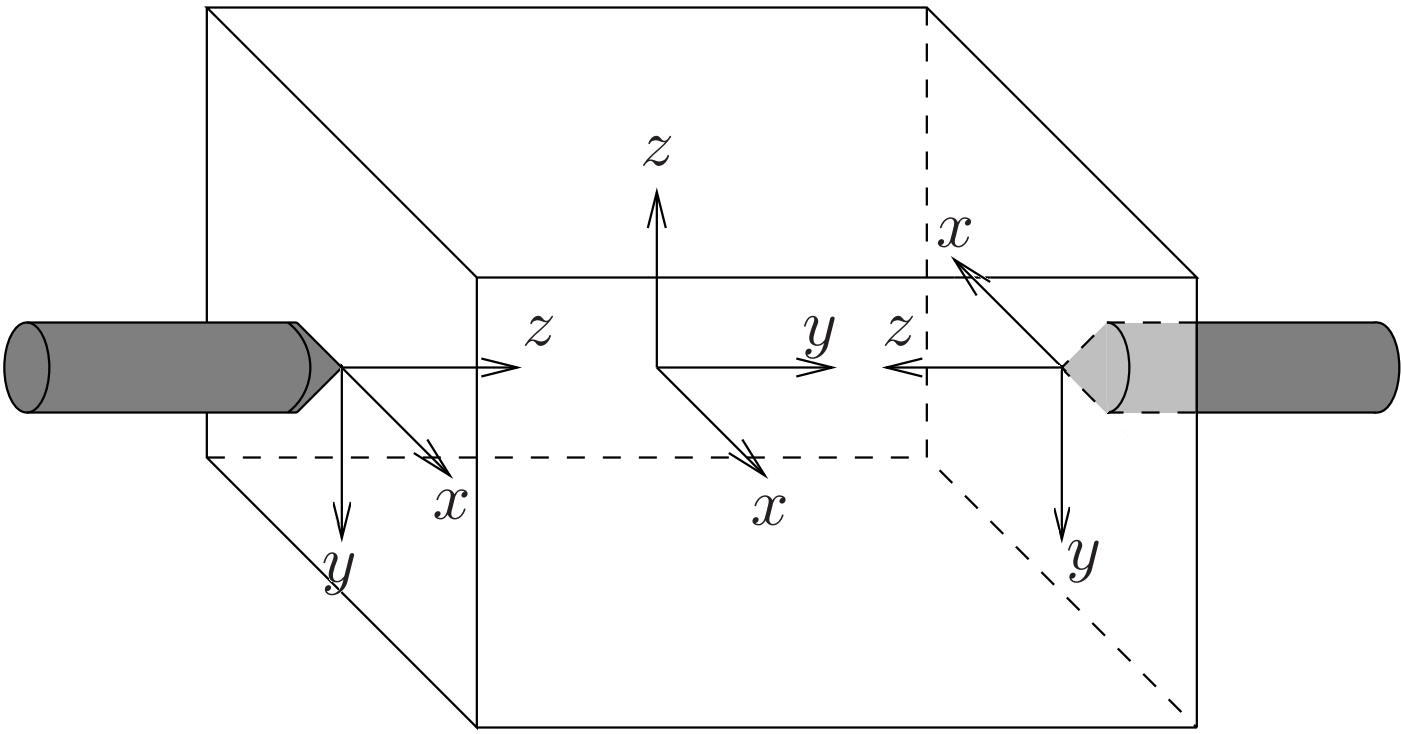}
	\caption{Spatial grasping considered for simulation.}
	\label{fig:SF}
\end{figure}

Here, we present the simulation results for the problem considered in Section \ref{sec:friction}. We consider the same simulation setup described in the previous subsection. 
The considered grasp can be seen in \figurename~\ref{fig:SF}. We assume that
the contacts are soft-finger contacts and that the $z$-axes of the contact coordinate frames point in the direction of the inward surface normals.
The line connecting the two contact points lies inside the friction cones, and hence according to \citet{nguyen1988constructing}, the grasp is force-closure. We have considered the coefficients of friction to be $\mu = 1$ and $\lambda = 1$, where $\mu = 1$ corresponds to a friction cone angle of 45\textdegree, i.e., the angle of the cone with respect to the surface normal. 
The interior of the friction cone for the $i$th contact has been considered as 
\begin{align}\label{eq:FCplanar_sim}
	\text{int}&(FC_{\scriptscriptstyle C_i}) = \notag \\
	& \ \  \left\{ f \in \mathbb{R}^4 : \sqrt{f_x^2+f_y^2} \leq \mu f_z - \delta_1, f_z \geq 0, \left| t_z \right| \leq \gamma f_z -\delta_2 \right\},
\end{align}
where $\delta_1$ and $\delta_2$ are small positive numbers and $f=\left [ f_x , f_y , f_z , t_z \right ]^T$.
For this simulation, we have chosen $\delta_1 = \delta_2 = 0.5$.

\begin{figure}[!t]
	\centering
	\includegraphics[width=11cm]{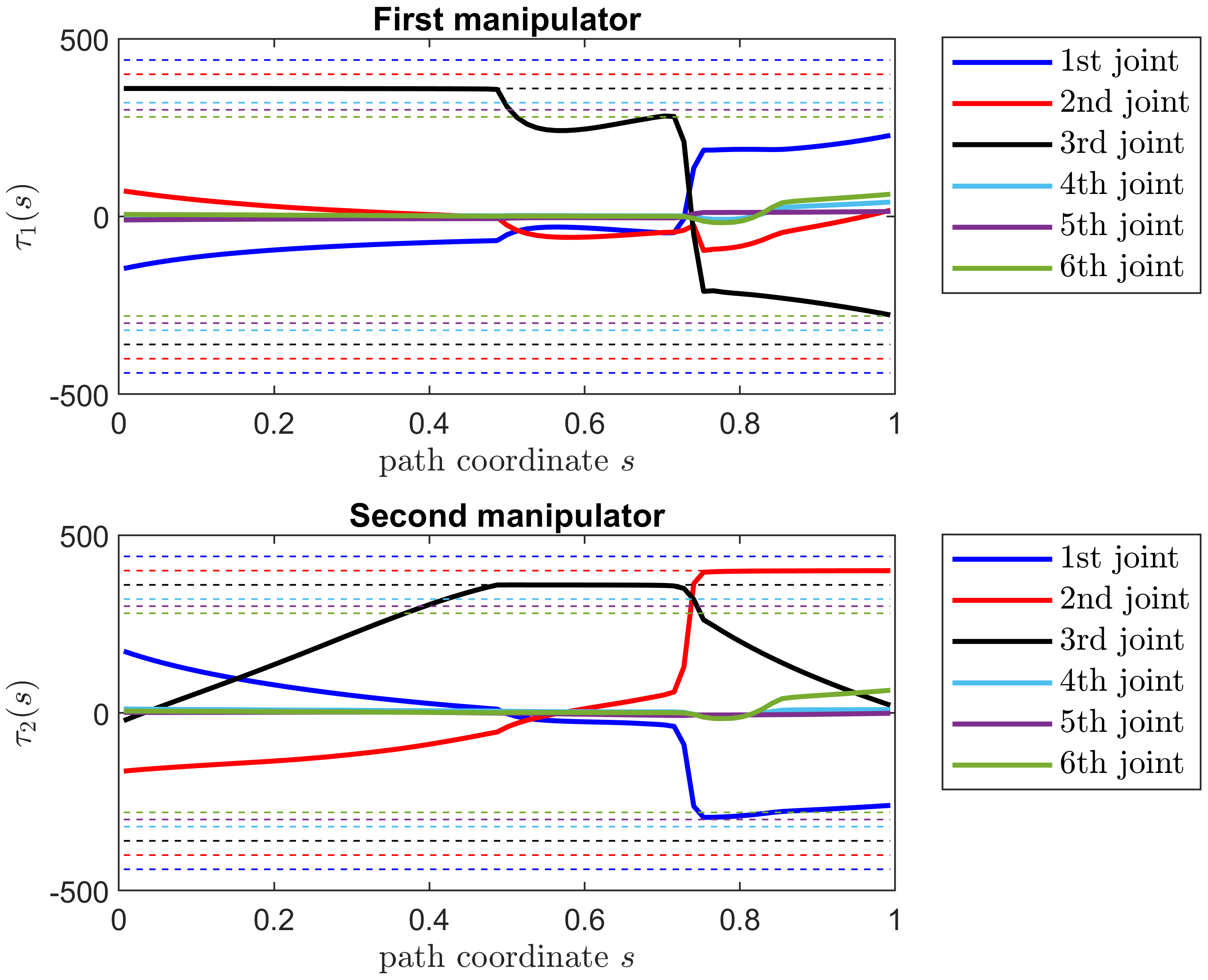}
	\caption{Joint torques/forces $\tau_{1}(s)$ and $\tau_{2}(s)$ for the case of contacts with friction. The units are Nm for a revolute and N for a prismatic joint.}
	\label{fig:torque_fc}
\end{figure}

\begin{figure}[!t]
	\centering
	\includegraphics[width=11cm]{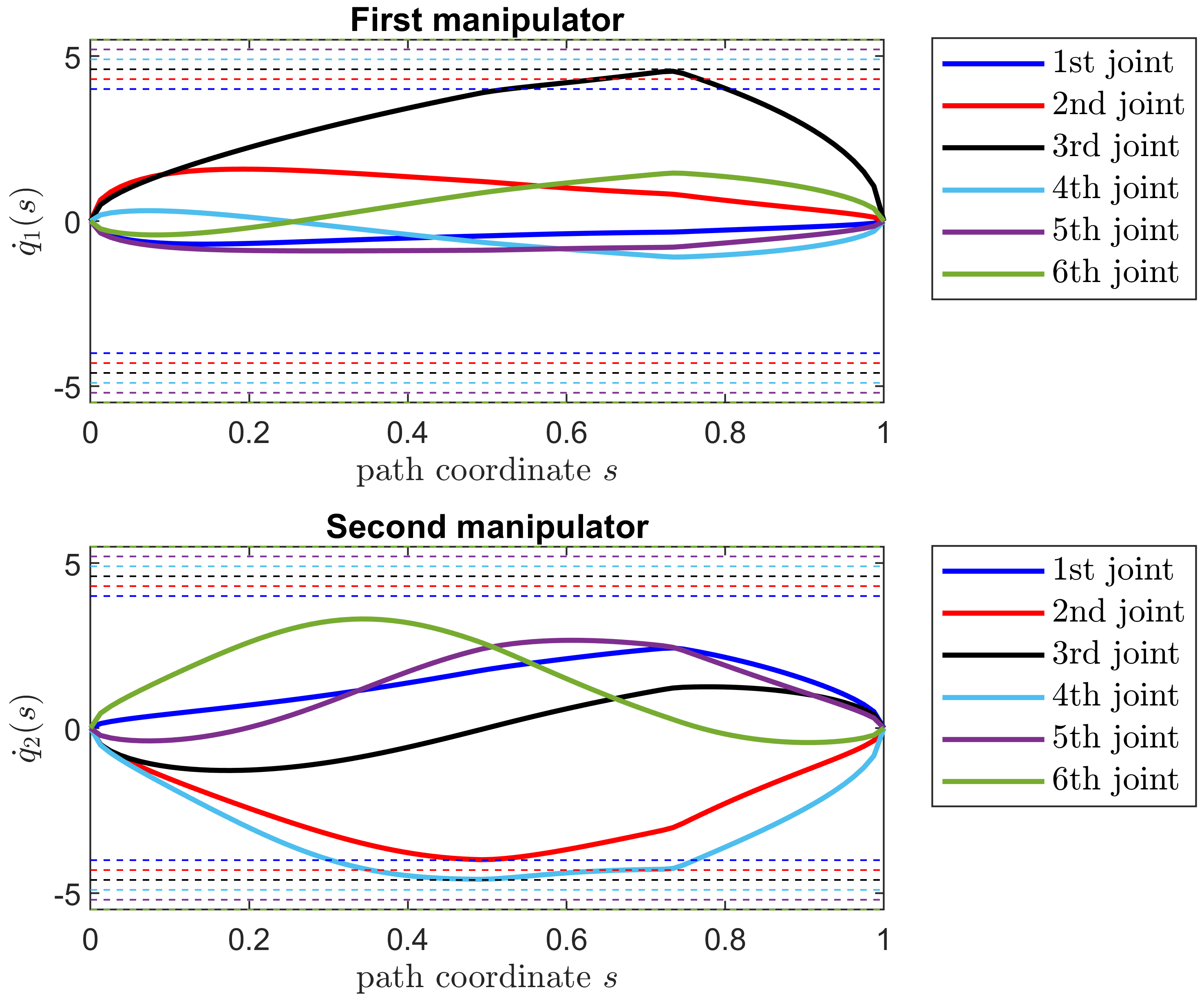}
	\caption{Joint velocities $\dot{q}_{1}(s)$ and $\dot{q}_{2}(s)$ for the case of contacts with friction. The units are rad/s for a revolute and m/s for a prismatic joint.}
	\label{fig:velocity_fc}
\end{figure}

\figurename~\ref{fig:torque_fc} and \ref{fig:velocity_fc} show the resulting joint torques/forces and joint velocities as functions of the path coordinate $s$, together with the considered bounds on them for both manipulators, respectively. The bounds are depicted in dashed lines.
Similar to the results from the case with rigid contacts, it can be seen that for every point along the path, either one of the joint torques or forces of the first or second manipulator is in saturation, or one of the joint velocities has reached its bound.

The simulation resulted in a minimal traversal time of 0.567 sec, which is greater that the minimal traversal time of the rigid contacts case. This is because of the additional constraints associated with friction cones that the contact forces must respect. 
Comparison between the minimal traversal times of rigid and frictional contact cases for other paths and orientations of the object can be found in Table~\ref{tb:comparison}. It can be seen that in all cases, the result obtained for rigid contacts is smaller than the one for frictional contacts.

\begin{figure}[!t]
	\centering
	\includegraphics[width=9.2cm]{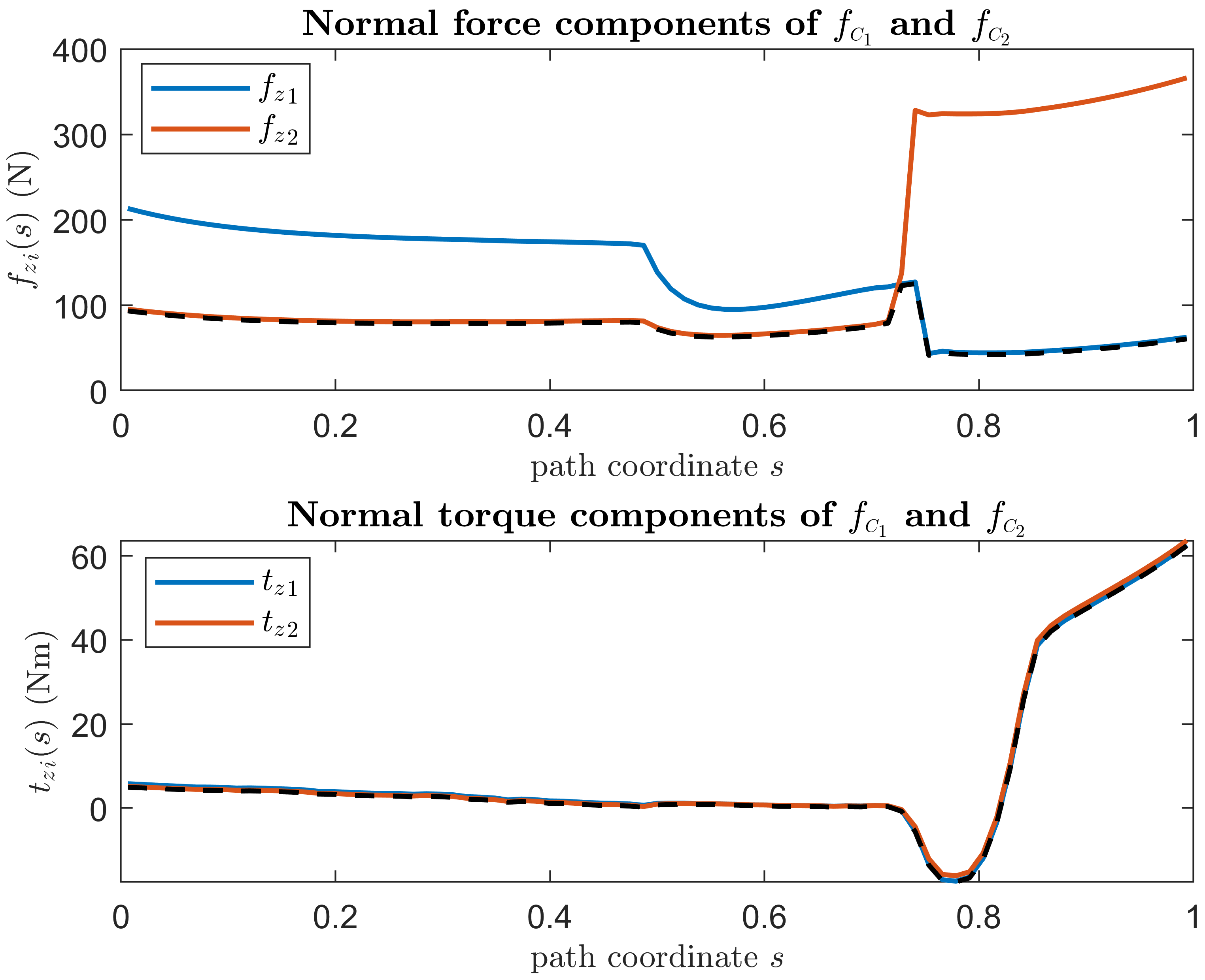}
	\caption{Normal components of contact wrenches $f_{\scriptscriptstyle C_i}, i \in \mathcal{N}$. The force and the torque components for the $i$th contact are denoted by ${f_z}_i$ and ${t_z}_i$, respectively. Black dashed lines show the internal forces.}
	\label{fig:normalTangential}
\end{figure}

\begin{figure}[!t]
	\centering
	\includegraphics[width=9.2cm]{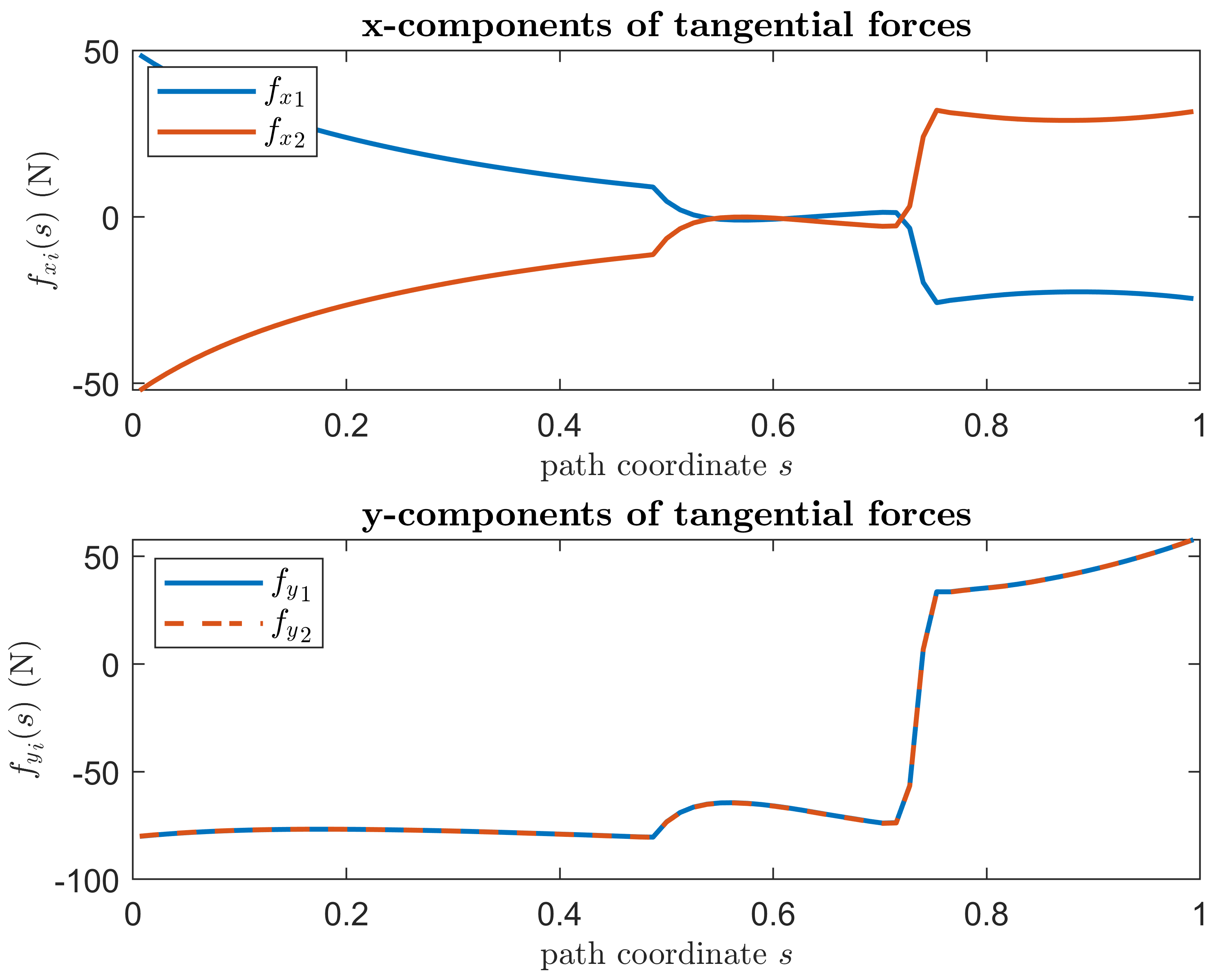}
	\caption{Tangential components of contact wrenches $f_{\scriptscriptstyle C_i}, i \in \mathcal{N}$. For the $i$th contact, the $x$-components are denoted by ${f_x}_i$ and the $y$-components are denoted by ${f_y}_i$.}
	\label{fig:normalTangential2}
\end{figure}

As mentioned earlier, in order to be able to firmly grasp an object, it is desirable that internal forces are present during the motion.
For the spatial grasp of \figurename~\ref{fig:SF}, internal forces are in the form of $f_{\scriptscriptstyle I_1} = \left [ 0 , 0 , \alpha , \beta \right ]^T$ and $f_{\scriptscriptstyle I_2} = \left [ 0 , 0 , \alpha , \beta \right ]^T$, i.e., the force or the torque components of contact wrenches in the normal direction with equal magnitudes. For these wrenches to be in the interior of the friction cone, $\alpha$ must be positive, meaning that the forces in the normal direction are applied in the positive direction of $z$-axis at each contact point. In order to investigate the presence of internal forces in the obtained solution from the simulation, the components of the wrenches $f_{\scriptscriptstyle C_1}$ and $f_{\scriptscriptstyle C_2}$ in the normal direction are shown in \figurename~\ref{fig:normalTangential}. 
The force components for both contacts are shown in the upper plot and the torque components can be seen in the lower plot.
Internal forces are the force or the torque components in the normal directions with equal magnitudes.
That is, the least of the two values of normal components for the forces. 
This is shown by the black dashed line in the upper plot in \figurename~\ref{fig:normalTangential}. 
Note that the obtained solution for the normal force components must be positive at every point along the path, since the wrenches must be in the interior of the friction cones. This is the case for the obtained solution and can be seen from the upper plot in \figurename~\ref{fig:normalTangential}. For the torque components, if at a point along the path both torques have the same sign, then the least of the magnitudes of the two torques, will be part of the internal force. This can be seen by the black dashed line in the lower plot.
The tangential components of the contact wrenches are plotted in \figurename~\ref{fig:normalTangential2}.
This simulation result shows that the solution of the proposed optimization problem involves internal forces belonging to the interior of the friction cone during the whole motion, which is a desirable property since it ensures a firm grasp.

\section{Conclusion}\label{sec:conclusion}
We have formulated the time-optimal path tracking problem for a multi-manipulator system carrying an object as a convex optimization problem. Two types of contact were considered: rigid contacts and contacts with friction. Numerical simulations were carried out for both cases. From the simulation results in both cases, it was seen that for every point along the path, either one of the joint torques or forces of the first or second manipulator was in saturation, or one of the joint velocities had reached its bound. This is in line with the results that can be found in the literature about the structure of the minimum-time control for multiple cooperating manipulators.
The resulting minimal traversal time from our approach was compared to the simulation results from using a particular wrench distribution for different paths, where the wrench distributions were chosen from other studies. For each path, our approach resulted in a smaller minimal traversal time, since it considers all possible combinations of wrenches and exploits the full actuation available to the system.
The case with contacts with friction resulted in a greater minimal traversal time compared to the case with rigid contacts. This is because of the additional constraints associated with friction cones that the contact wrenches must respect in the case of contacts with friction. 
Furthermore, in the same case it was seen that the solution of the proposed optimization problem involves internal forces that lie in the interior of the friction cone during the whole motion, which is a desirable property for the purpose of having a firm grasp. In future work we plan to study the extension to mobile manipulators, where manipulators are mounted on a mobile platform. Extension to consider rolling contacts where grasping involves moving rather than fixed contact points is another possibility. 



\section*{Declaration of competing interest}
The authors declare that they have no known competing financial interests or personal relationships that could have appeared to influence the work reported in this paper.

\bibliography{myrefs}
\bibliographystyle{elsarticle-harv}

\vfill

\end{document}